# A Mutual-Structure Weighted Sub-Pixel Multimodal Optical Remote Sensing Image Matching Method

Tao Huang[a], Hongbo Pan[a]*, Nanxi Zhou[a], Siyuan Zou[a], Shun Zhou[a]

[a]*School of Geosciences and Info-Physics, Central South University. #932 Lunan Road, Changsha, Hunan Province, 410083, P.R. China.*

*Corresponding author: Pan, Hongbo (hongbopan@csu.edu.cn)

# A Mutual-Structure Weighted Sub-Pixel Multimodal Optical Remote Sensing Image Matching Method

Sub-pixel matching of multimodal optical images is a critical step in combined application of multiple sensors. However structural noise and inconsistencies arising from variations in multimodal image responses usually limit the accuracy of matching. Phase congruency mutual-structure weighted least absolute deviation (PCWLAD) is developed as a coarse-to-fine framework. In the coarse matching stage, we preserve the complete structure and use an enhanced cross-modal similarity criterion to mitigate structural information loss by PC noise filtering. In the fine matching stage, a mutual-structure filtering and weighted least absolute deviation-based is introduced to enhance inter-modal structural consistency and accurately estimate sub-pixel displacements adaptively. Experiments on three multimodal datasets-Landsat visible-infrared, short-range visible-near-infrared, and UAV optical image pairs demonstrate that PCWLAD consistently outperforms eight state-of-the-art methods, achieving an average matching accuracy of approximately 0.4 pixels. The software and datasets are publicly available at https://github.com/huangtaocsu/PCWLAD.

Keywords: multimodal matching; optical images; phase congruency; accuracy; mutual structure

## 1.Introduction

The combined application of multimodal images is becoming increasingly vital in remote sensing. Optical images with different spectral ranges can capture diverse types of information and are widely applied in areas such as target detection (Jiang et al. 2024), environmental monitoring (Huang et al. 2021), and geometry calibration (Zhao et al. 2024). Optical imaging sensors, commonly deployed on satellite and UAV platforms, typically include the visible, near-infrared, and short-wave infrared bands to capture ground object reflectance and thermal characteristics (Wulder et al. 2022). Visible to near infrared (VNIR) images typically have higher resolution and provide detailed texture and variations in light and shadow, but are highly sensitive to weather conditions (Zhao and Li 2021). However, infrared images are resistant to these interferences and reveal features that are challenging to detect with VNIR, but they have lower resolution and provide less texture information (Hu et al. 2024). Therefore, the high-precision fusion (Ma, Ma, and Li 2019) of VINR and infrared images enables complementary advantages, improving performance in applications such as infrastructure defect inspection (Motayyeb et al. 2023), precision agriculture (Maimaitijiang et al. 2020), and image recognition and classification (He et al. 2021).

Significant nonlinear radiometric and geometric deformation differences in multimodal image data often reduce the accuracy of matching. The correlation between multimodal optical images decreases due to differences in pixel intensity, gradient changes, and texture features. As a result, least-squares matching methods (Gruen 1985; Sedaghat and Ebadi 2015) that combine geometric and radiometric features often fail and require accurate initial values (Shin and Muller 2012). To enhance the accuracy of multimodal matching, current mainstream methods typically employ a fine-tuning process after coarse matching (Hou, Liu, and Zhang 2024; Fan et al. 2022).However, this common strategy often ignores structural consistency and noise. First, the varying response of different modal remote sensing images often produce numerous unique local or detailed texture features (Liao et al. 2024). This non-repeating texture creates significant differences in structural consistency, introducing noise into high-precision matching. Second, processing PC noise (Donoho 1995) with the same parameters and methods across different modes introduces significant noise, reducing similarity and weakening accuracy optimization. Multimodal matching accuracy remains limited due to the intrinsic structured noise within the PC and the influence of nonlinear phase distortions.

The main contributions of the proposed PCWLAD method are summarized as follows:

(1) Considering that noise filtering of PC features can eliminate rich structural information, we preserve the full structural representation to enhance the similarity between multimodal optical images. Furthermore, we employ the enhanced cross-modal similarity measure to improve the robustness of the coarse matching process.

(2) To mitigate the adverse effects of structural inconsistencies and noise on sub-pixel estimation, we introduce mutual-structure weighting to reinforce structural consistency and incorporate a robust WLAD strategy to improve resistance to noise, thereby achieving reliable and accurate sub-pixel matching in the fine matching stage.

## 2.Related Studies

Existing multimodal optical image matching methods fall into three categories: area-based, feature-based, and deep learning-based methods. The three methods enhance the robustness of multimodal matching, yet the matching accuracy remains limited, making them difficult to apply in higher precision applications like camera calibration.

### *2.1. Area-based methods*

Area-based matching typically uses a metric to assess the similarity of template windows. Directly utilizing the similarity of image intensity such as sum of squared differences (SSD) and normalized cross-correlation (NCC) (Zitová and Flusser 2003; Yoo and Han 2009) , are generally difficult to apply in remote sensing multimodal image matching. On the other hand, mutual information (MI), which uses a joint probability distribution histogram to assess similarity (Maes et al. 1997), can handle nonlinear radiometric differences and has been applied to multimodal images in remote sensing (Suri and Reinartz 2010). However, MI disregards structural image details, its ability to capture cross-modal similarity diminishes.

Structural features have been incorporated as a similarity measure for multimodal matching (Ye et al. 2017), enhancing its robustness to nonlinear intensity differences. Some dense matching studies have proved that template matching based on structural and shape information can create similarity maps of pixel-level features and suppress the effects of nonlinear radiation distortion (Revaud et al. 2016). Similarity assessments in these methods include local self-similarity (LSS) (Shechtman and Irani 2007) and histogram of oriented gradients (HOG) (Dalal and Triggs 2005).These structural features, based on a relatively sparse sampling grid, fail to accurately capture the structural features of the image (Ye et al. 2019). Ye et al. (2017) introduced PC maps to construct histogram of oriented phase congruency (HOPC) for accurately capturing the structure of images and defined the channel features of oriented gradients (CFOG) (Ye et al. 2019) for fast similarity measurement based on the Fast Fourier Transform (FFT) in the frequency domain. Meng et al. combined template matching with weights, multilevel local max-pooling, and max index backtracking to enhance the matching accuracy between UAV visible and infrared images (Meng et al. 2022). The accuracy of these template matching methods depends on window size, structural noise, and the consistency of structural response, with accuracy typically limited to the pixel-level.

### *2.2. Feature-based methods*

Unlike area-based matching methods, feature-based methods are more robust and efficient in handling geometric deformations; however, they face challenges related to feature detection repeatability and the robustness of descriptor similarity in multimodal images (Jiang et al. 2021). Differences in pixel intensity and gradient distribution between multimodal images make traditional methods like SIFT (Lowe 2004) ineffective, prompting the development of improved SIFT-based approaches to address these challenges. These improvements increase robustness against noise (Yu et al. 2021), gradient optimization (Xiang, Wang, and You 2018), and image discrepancies by refining feature extraction and descriptor construction (Chang, Wu, and Chiang 2019). However, multimodal gradient changes involve nonlinear distortions, making gradient-based enhancements unreliable.

In the multimodal feature detection stage, current work primarily focuses on detecting features on structured features, such as PC, to ensure repeatability (Kim et al. 2012; Nunes and Pádua 2024). RIFT (Li, Hu, and Ai 2019) detects FAST (Viswanathan 2009) features using PC maximum moment and makes descriptors through a multi-scale, end-to-end loop structure to achieve rotation invariance. Histogram of absolute phase congruency gradients (HAPCG) (Yao et al. 2021) employs anisotropic filtering to compute PC and uses the HARRIS (Harris and Stephens 1988) for feature detection. However, the matching accuracy of these methods is often constrained by the joint feature detection performance on PC maps (Liao et al. 2024). To further improve the uniqueness and rotation invariance of descriptors, Fan et al. introduced the OFM (Fan et al. 2023) method to construct a rotation-invariant descriptor with oriented filter. Sparse sampling description-based matching method (SSDM) (Fan, Pi, and Wang 2025) proposed a hierarchical matching strategy driven by a global transformation model, which substantially enhances the matching accuracy of descriptors (Zheng et al. 2025). Most feature descriptors are robust to nonlinear radiometric differences and effectively match multimodal images. However, PC-based detection methods are constrained by the same structural limitations as area-based methods.

### *2.3. Deep learning-based methods*

Data-driven deep learning can harness high-level semantic information from large volumes of training data (Jiang et al. 2021), extracting more refined common features than traditional handcrafted methods. It is better to adapt to geometric and radiometric differences, as well as noise, between multimodal remote sensing images (Li et al. 2025).These methods fall into two categories: one replaces a key step in the traditional matching pipeline with a single-stage neural network, while the other uses an end-to-end model to replace the entire image-matching process.

Single-stage networks capitalize on the strengths of fully data-driven approaches and high-dimensional deep features (Liu and Deng 2015; Oquab et al. 2023) by integrating deep networks into specific stages of matching (Uss et al. 2022; Yang, Dan, and Yang 2018). Area-based matching methods integrate a specific network by replacing intensity information or shallow structural features, then evaluating the similarity between the output feature maps (Merkle et al. 2017; Zhang et al. 2022). These methods enhance resistance to nonlinear radiometric differences and improve cross-modal similarity (Nguyen et al. 2018; Zhang et al. 2021; Ye et al. 2024). In the feature matching stage, many studies utilize specific networks to construct robust descriptors (Quan et al. 2022; Ye et al. 2018), while fewer focus on high-precision multimodal feature detection. To enhance the repeatability of feature detection, some researchers have leveraged the global attention mechanism to generate more discriminative descriptors for each feature point (Cui et al. 2022; Liang et al. 2023; Li, Han, and Ye 2022). SuperGlue (Sarlin et al. 2020) reformulates the feature matching task as a differentiable optimization problem, enabling its solution using graph neural networks (Lindenberger, Sarlin, and Pollefeys 2023). These descriptors improve the overall performance of multimodal feature matching, but neglect the accuracy of feature detection.

End-to-end deep networks directly predict the underlying transformation model and typically adopt a fully automated, multi-scale, multimodal image matching framework (Hughes et al. 2020). LoFTR (Sun et al. 2021) represents a significant advancement in this field, achieving reliable matching even in low-texture regions. Faced with extreme variations in real-world images, RoMa (Edstedt et al. 2024) integrates specialized convolutional neural network features by building a fine feature pyramid and employing a custom transformer-based matching decoder to establish dense and robust matching. Another type of end-to-end approach involves modality unification through style transfer (Aggarwal, Mittal, and Battineni 2021). The effectiveness of image modality unification depends on training sample size and scene diversity, leading to variations in modality conversion that affect registration accuracy.

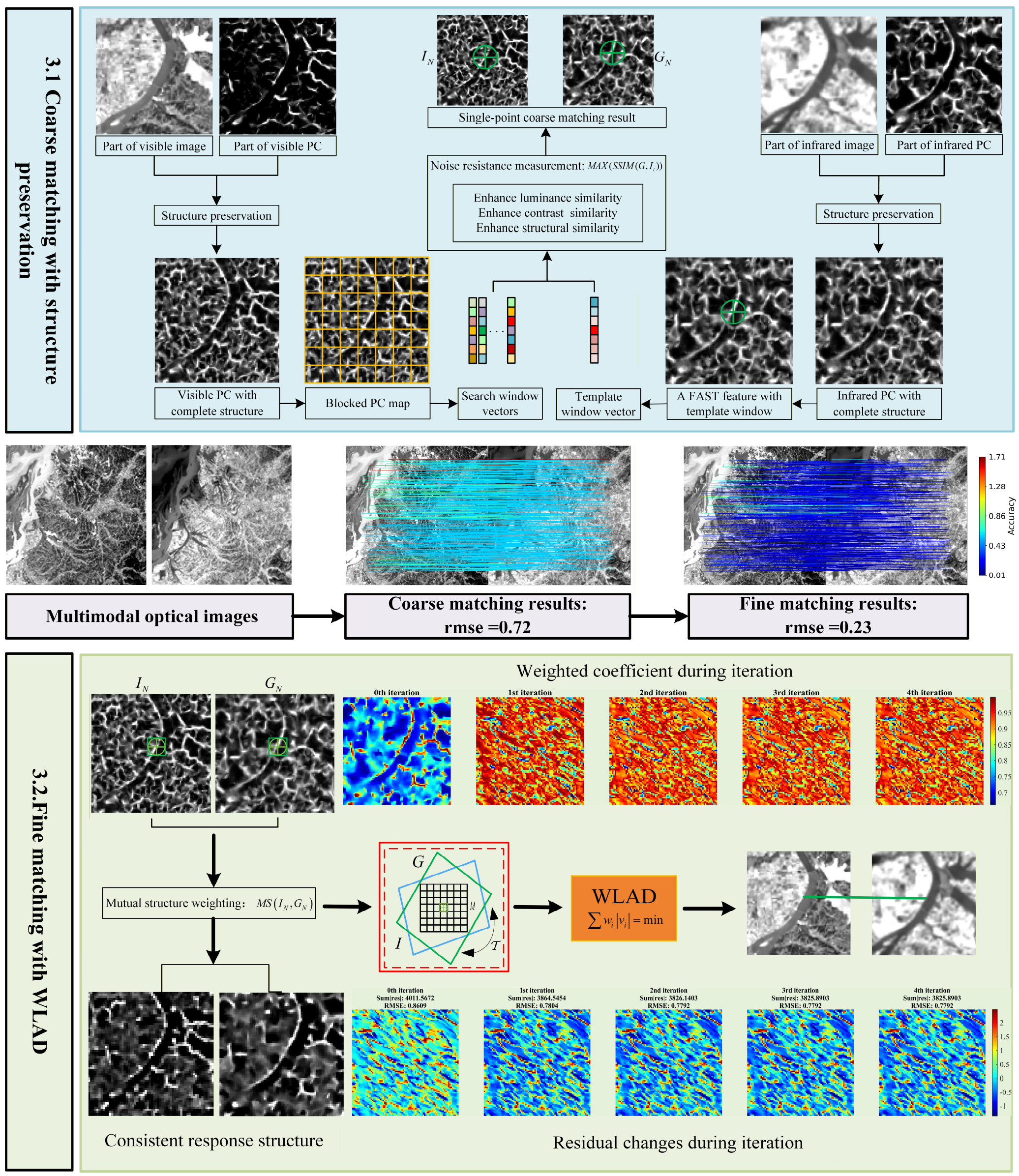

Figure 1. Pipeline of the PCWLAD method.

## 3.Methodology

We propose a high-accuracy multimodal matching method for optical images, consisting of two main steps. First, we compute PC without noise filtering to preserve structural information, then detect FAST features on the visible PC image, which contains richer texture. A template window is selected around each feature point, and a corresponding search window is constructed on the thermal infrared PC map. Enhanced SSIM is used for coarse matching (Wang and Bovik 2009). Second, we estimate geometric and radiometric transformations between coarsely matched windows, apply mutual-structure weighting (Shen et al. 2015) to reinforce structural consistency, and use the WLAD strategy to suppress noise and refine correspondences, achieving sub-pixel accuracy. A parameter-adaptive method (Huang, Pan, and Zhou 2024) removes matching outliers.

### *3.1. Coarse matching with structure preservation*

PC is widely used for feature matching in multimodal images due to its invariance to illumination and contrast, making it a robust choice for image analysis under varying imaging conditions. Kovesi refined the computational model of PC by employing Log-Gabor wavelets across multiple scales and orientations (Kovesi 1999), enhancing its robustness and accuracy:

$$PC(x,y)=\frac{\sum_{o}\sum_{n}W_o(x,y)A_{no}(x,y)\Delta\Phi_{no}(x,y)-T_o}{\sum_{o}\sum_{n}A_{no}(x,y)+\varepsilon} \tag{1}$$

where $(x,y)$ denote the coordinates of a point in the image, $W_o(x,y)$ is a weighting factor for a given frequency distribution, $A_{no}(x,y)$ represents the amplitude at for wavelet scale $n$ and orientation $o$, $T_0$ is the noise threshold, and $\varepsilon$ is a small constant to prevent division by zero. $\Delta\Phi_{no}(x,y)$ is a more sensitive definition of phase difference.

For the noise $T_0$, it can be defined as:

$$T_0 = \mu_R + k * \sigma_R \tag{2}$$

where $k$ is typically in the range 2 to 3, $\mu_R$ is the mean of the Rayleigh distribution, $\sigma_G$ represents the standard deviation of the Gaussian distribution describing the position of the total energy vector.

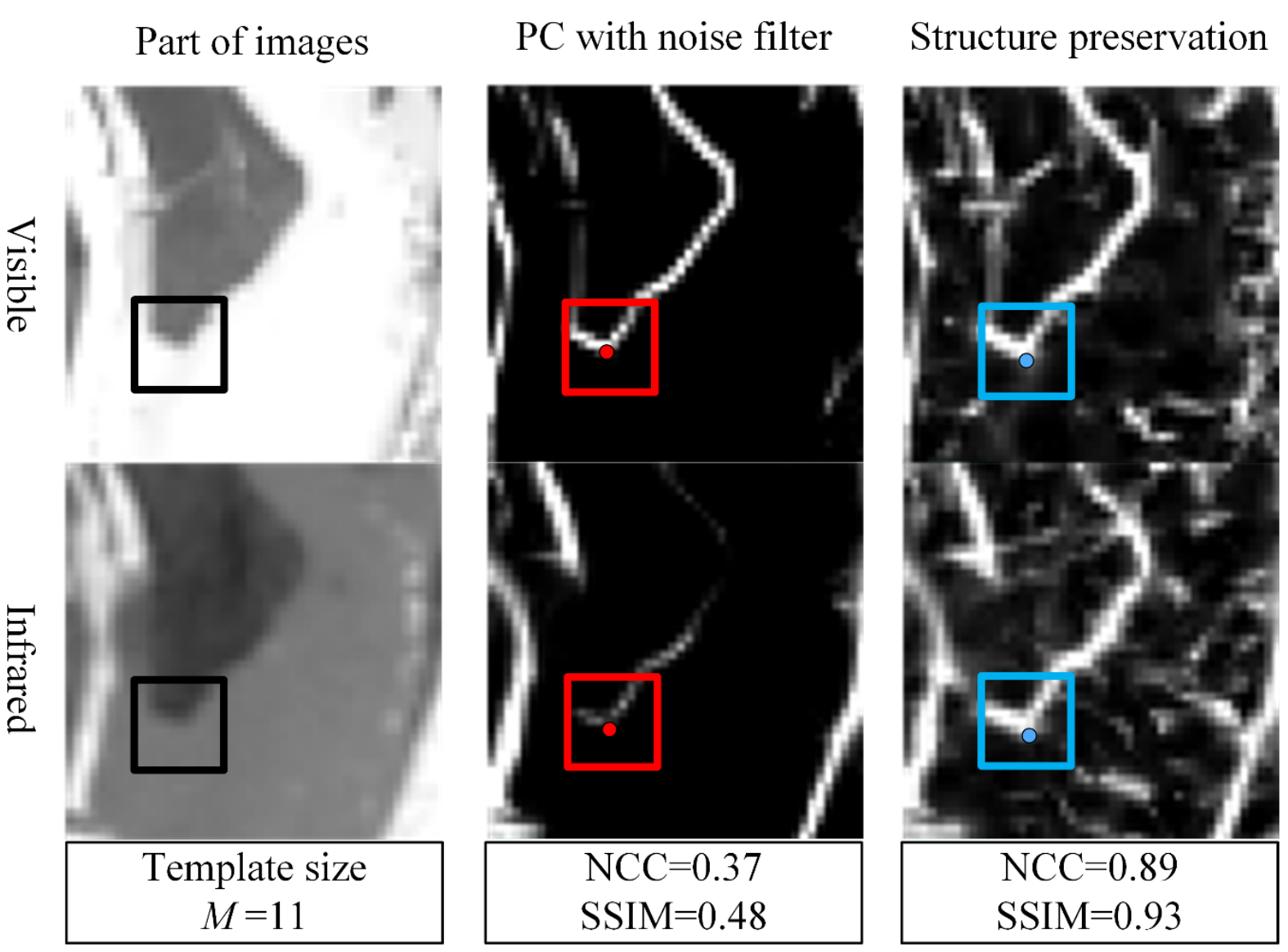

Figure 2. The impact of noise filter in PCs on structural integrity and similarity.

In practical calculations, the mean and standard deviation of the Rayleigh distribution is estimated based on the response amplitudes across all positions in a given direction and multiple scales. This estimation can lead to excessive or insufficient noise suppression, potentially distorting local texture structures. As shown in Figure. 2, their structural consistency becomes challenging after the visible and infrared local PC maps are processed using the same noise method. However, the PC map without a noise filter, although affected by noise, preserves the continuity and stability of edge textures relative to the original image, particularly in regions with prominent texture. Therefore, we retain the complete structural information to enhance the structural similarity between multimodal images, while handling noise during the matching process.

We apply FAST feature detection to the phase congruence map of the visible image without noise reduction. Utilizing the structural similarity of PC, we construct a square template window for each feature in the visible PC map with a neighborhood size of $M$, representing the phase values of all cells within the window as a vector $G$. In Figure.1, a search window of the same size is constructed on the PC map of the thermal infrared image, using the phase values within the window to form a vector $I$. The sliding window method is then employed to identify the template and search windows with the highest SSIM coefficient, achieving coarse matching.

Considering the nonlinear radiometric differences between visible and infrared images, the PC map exhibits noise in the form of inconsistent texture structure responses. Additionally, the PC maps of the two modalities contain nonlinear components induced by the differences in radiation characteristics. The local SSIM index quantifies similarity based on three key elements: luminance similarity $L(G,I)$, which measures the consistency of brightness values between patches; contrast similarity $C(G,I)$, which evaluates variations in intensity. structural similarity $S(G,I)$, which captures the correlation of local structural patterns. To enhance similarity, the weights are set as follows:

$$SSIM(G,I) = [L(G,I)]^{\alpha} \cdot [C(G,I)]^{\beta} \cdot [S(G,I)]^{\gamma} \tag{3}$$

$\alpha$, $\beta$ and $\gamma$ are weights for luminance, contrast, and structure components, respectively. To enhance similarity, the weights are set as $\alpha = 0.5, \beta = 0.8, \gamma = 1.3$ .For luminance similarity, contrast similarity, and structural similarity, the definition is as follows:

$$\begin{cases} L(G,I) = \dfrac{2\mu_G\mu_I + C_1}{\mu_G^2 + \mu_I^2 + C_1} \\ C(G,I) = \dfrac{2\sigma_G\sigma_I + C_2}{\sigma_G^2 + \sigma_I^2 + C_2} \\ S(G,I) = \dfrac{\sigma_{GI} + C_3}{\sigma_G\sigma_I + C_3} \end{cases} \tag{4}$$

Where $\mu_G$ and $\mu_I$ are the means of the $G$ and $I$ vectors respectively, $\sigma_G$ and $\sigma_I$ are the standard deviations of the $G$ and $I$ vectors respectively, $\sigma_{GI}$ is the covariance of the $G$ and $I$ vectors. $C_1$, $C_2$ and $C_3$ are stabilization constants that prevent numerical instability when both $\mu_G^2 + \mu_I^2$ and $\sigma_G^2 + \sigma_I^2$ approach zero, ensuring reliable SSIM computations even in regions with low contrast or uniform intensity.

### *3.2. Fine matching with WLAD*

Let $G(p)$ and $I(q)$ denote the PC responses in the reference and target images, respectively. Consider an $M \times M$ neighborhood patch $\Omega$ centered at a coarse match. We enumerate the pixels in $\Omega$ as $\Omega = \{p_i, i = 1, \cdots, N\}$, where $N = M^2$ and $p_i \in \mathbb{R}^2$ is the 2D pixel coordinate in the reference patch. For each $p \in \Omega$, the corresponding location in the target patch is $q = \mathcal{T}(p)$. The value $I(\mathcal{T}(p))$ is evaluated via interpolation when $\mathcal{T}(p)$ is non-integer. To account for both radiometric differences and geometric distortions, we adopt a local linear radiometric model together with a local geometric mapping:

$$G(p) = \beta_0 + \beta_1 \cdot I(\mathcal{T}(p)) + e(p) \tag{5}$$

Here, $\mathcal{T}(p)$ maps reference coordinates to target coordinates. Parameters $\beta_0$ and $\beta_1$ represent radiometric offset and gain, respectively. The residual $e(p)$ aggregates model errors, noise, and structural mismatches.

This approach introduces a novel dual-adaptive mechanism that combines structural priors with residual-based adaptation. Let $b_i$ be the design vector and $l_i$ be the constant term for the i-th observation. The estimation problem is formulated as follows:

$$\hat{X} = \arg\min_{X} \sum_{i=1}^{N} w_i \left| \mathbf{b}_i^T X - l_i \right| \tag{6}$$

Define the design matrix $B = [b_1, \cdots, b_N]$ and the observation vector $L = [l_1, \cdots, l_N]^T$. Let $W^{(t)} = diag[w_1^{(t)}, \cdots, w_N^{(t)}]$ be the diagonal weight matrix at iteration $t$. By integrating observations and weights into the WALD framework (Wang, Li, and Jiang 2007), we derive a robust estimator capable of handling structural mismatches. This is achieved through an iteratively reweighted least squares algorithm (Daubechies et al. 2010), with the adaptive weighting mechanism central to the process. At iteration $t$, the weighted least squares problem is solved as follows:

$$X^{(t+1)} = \left(B^T W^{(t)} B\right)^{-1} B^T W^{(t)} L \tag{7}$$

To incorporate structural priors that improve robustness when images come from different sensors, we jointly estimate a filtered target response $\hat{I}$ and a mutual-structure image $MS$. Mutual structure filtering defines the overall objective function:

$$E(\hat{I}, MS, \beta_0, \beta_1) = E_S(I, G, \beta_0, \beta_1) + E_d(I, G) + E_r(\beta_0, \beta_1) \tag{8}$$

Here, $E(\hat{I}, MS, \beta_0, \beta_1)$ is the total cost to be minimized. It is composed of three terms. $E_S(I, G, \beta_0, \beta_1)$ is the mutual-structure consistency term, which enforces structural agreement between $I$ and $G$ through the regression models parameterized. $E_d(I, G)$ is the data term, which constrains the estimated images $I$ and $G$ to remain reasonable. $E_r(\beta_0, \beta_1)$ is the regularization term, which penalizes unstable or overly complex regression coefficients $r_0$ and $r_1$, improving robustness and smoothing. The optimization is a process to get filtering output $I$ and mutual structure $MS$ from $I(q)$ and $G(p)$ after reasonable smoothing (Shen et al. 2015).

The mutual structure $MS$ provides prior weights that can suppress structural inconsistencies and guide robust iterative estimation. The weights are updated using the residuals $d_i^{(t)} = b_i^T X^{(t)} - l_i$ computed from the current parameter estimates:

$$w_i^{(t+1)} = \frac{1}{1 + \left|d_i^{(t)}\right| / \sigma_s} \tag{9}$$

Here, $\sigma_s$ is used to normalize the residual scale. This update rule implements a dual-adaptive scheme: the mutual structure weight provides structural prior confidence, while the residual $d_i^{(t)}$ dynamically adjusts the weight during the iteration based on the model fit, further suppressing the influence of PC outliers. Finally, the parameters are estimated using the least squares method, and the iterative process terminates when the parameter change falls below a predefined threshold.

## 4.Experiment and analysis

### *4.1. Datasets*

To assess the precision of optical multimodal images, three datasets were used, as shown in Figure 3 and detailed in Table 1.

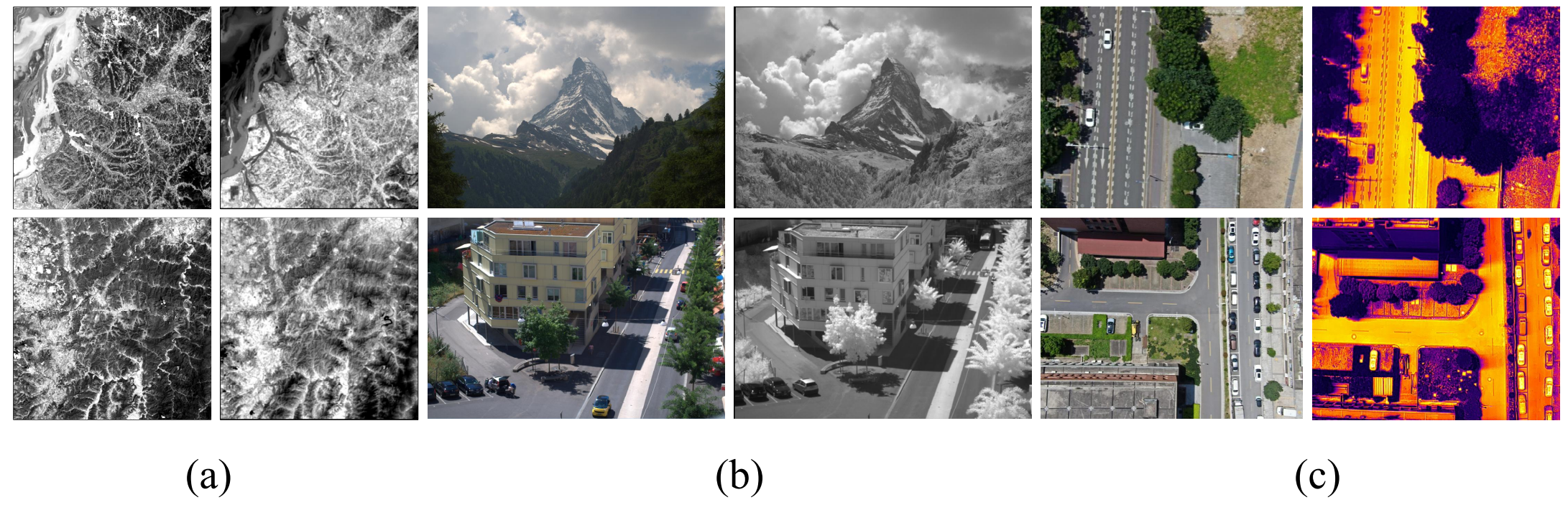

(a) (b) (c)

Figure 3. Examples of three types of experimental data. (a)Landsat data. (b)BGRNIR data. (c)UAV data.

Landsat Data: Offering registration accuracy better than 0.05 pixels (Tilton, Lin, and Tan 2017; Tilton et al. 2019), small image patches were extracted for test samples, with additional resampled images created using sub-pixel shifts. BGRNIR Dataset: This dataset consists of 477 registered RGB and NIR image pairs (Brown and Süsstrunk 2011), captured with Nikon D90 and Canon T1i cameras using B+W 486 (visible) and 093 (NIR) filters. UAV Dataset: Acquired using a DJI H20T camera for visible and infrared images, the data was collected near the School of Geosciences and Info-Physics at Central South University. Overlapping visible and thermal infrared images were grouped from different viewpoints. COLMAP (Schönberger and Frahm 2016) was used for bundle adjustment and image un-distortion, resulting in a reprojection error of approximately 0.3 pixels.

Table 1. Introduction of the datasets.

| NO. | Image type | Resolution(pixels) | GSD | Notes |
|---|---|---|---|---|
| 1 | Visible /Infrared | 512*512 | 100m | Images obtained from registered Landsat data. |
| 2 | Visible/Near-infrared | 1024*768 | 0.1m | Images captured from a registered close-up camera. |
| 3 | Visible /Infrared | 648 *516 | 0.08m | Images generated through undistorted UAV data. |

### *4.2. Evaluation criteria and parameter settings*

We present both qualitative and quantitative experimental results to fully demonstrate the proposed method. The quantitative evaluation uses four indicators. (1) NCM is defined as the number of correct matches. (2) CMR (correct match ratio) is calculated as CMR = NCM/C, where C is the total number of matched point pairs. (3) SR (success rate) represents the ratio of the number of extracted features to C or the ratio of successful convergence during fine matching. (4) RMSE (root mean square error) is calculated from matching point residuals, reflecting match accuracy. A smaller RMSE indicates higher accuracy. If the matching data has real coordinates, the fundamental matrix is unnecessary, and the corresponding epipolar line can be used for matching.

The accuracy of evaluation methods based on manual point selection is limited by the precision of the points and the inherent deformation of images (Ye et al. 2017). In near-vertical remote sensing images, terrain-induced projection distortions are minimal, making affine approximation feasible. However, in more complex scenarios with perspective distortion and terrain-induced projection differences, the fundamental matrix offers a more accurate estimation, describing the epipolar geometry between two images (Barath et al. 2023). For a 3D point in the same scene, its projections in the two images are $X_1$ and $X_2$, respectively. These projections satisfy the following relationship: ${X_2}^T H X_1 = 0$. $H$ can be estimated by inputting the matching point coordinates into the RANSAC (Fischler and Bolles 1981) algorithm, with the reprojection error threshold set to 2 pixels.

For parameter settings, we extracted 1000 FAST features from the visible PC map. Based on SR and RMSE performance across all datasets, the coarse matching template size $M$ is set to 101 (Section 4.3), and the fine matching template size is set to 81 (Section 4.4) to optimize matching accuracy. The maximum number of iterations is set to 20 during the precision optimization process. The iteration terminates when the offset corrections are less than 0.05 pixels, and the SSIM coefficient exceeds both 0.4 and the initial SSIM coefficient.

### *4.3. Analysis and evaluation of coarse matching*

We evaluated our proposed method alongside the HOPC algorithm using various window sizes, applying the PC and NCC metrics, which do not fully preserve structural information. We extracted the same 200 features as shown in Figure 4 and tested CMR, SR, and RMSE under different window sizes, comparing them between our coarse matching method and HOPC.

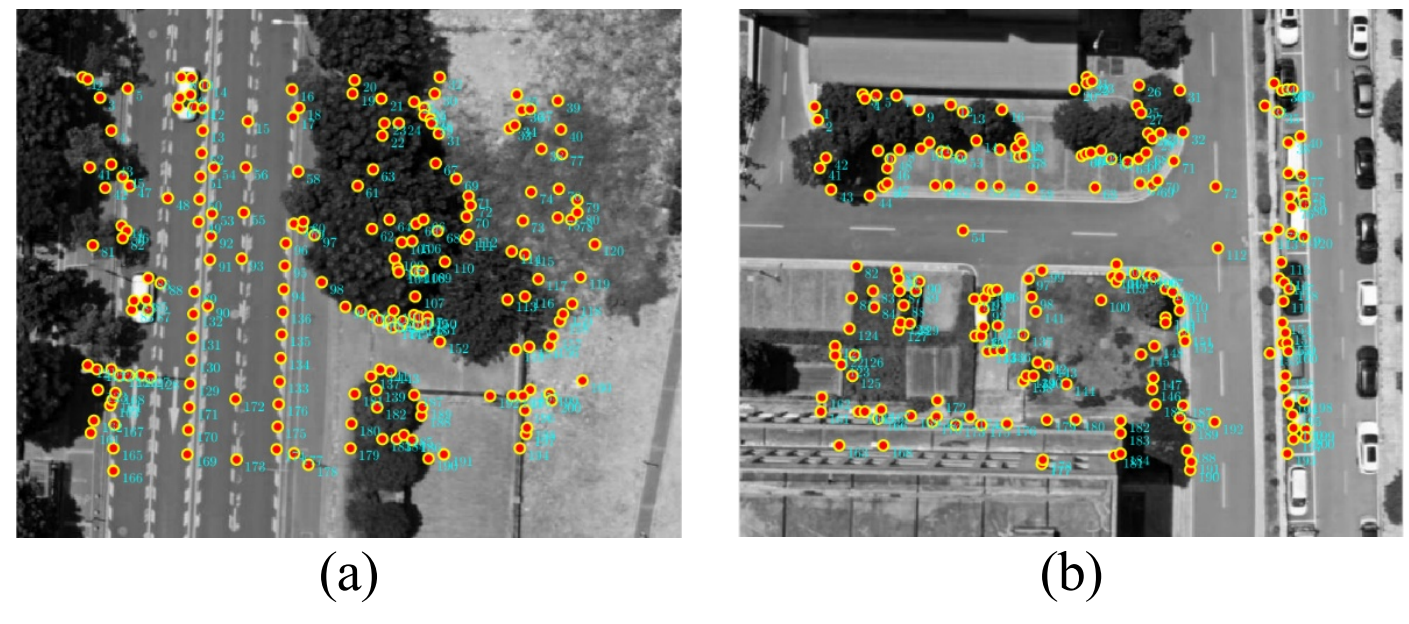

(a) (b)

Figure 4. The same 200 feature distributions. (a) Feature distribution of UAV data1. (b) Feature distribution of UAV data 2.

As shown in Figure 5(a), the SR increases with the increase of the window size in both methods. However, our coarse-matching method outperforms the HOPC method in terms of success rate (SR), achieving approximately twice the SR. As shown in Table 2, the average SR across different window sizes is 31% higher for our method. This improvement is primarily attributed to the incorporation of the complete PC structural information and the use of more robust similarity measurement criteria.

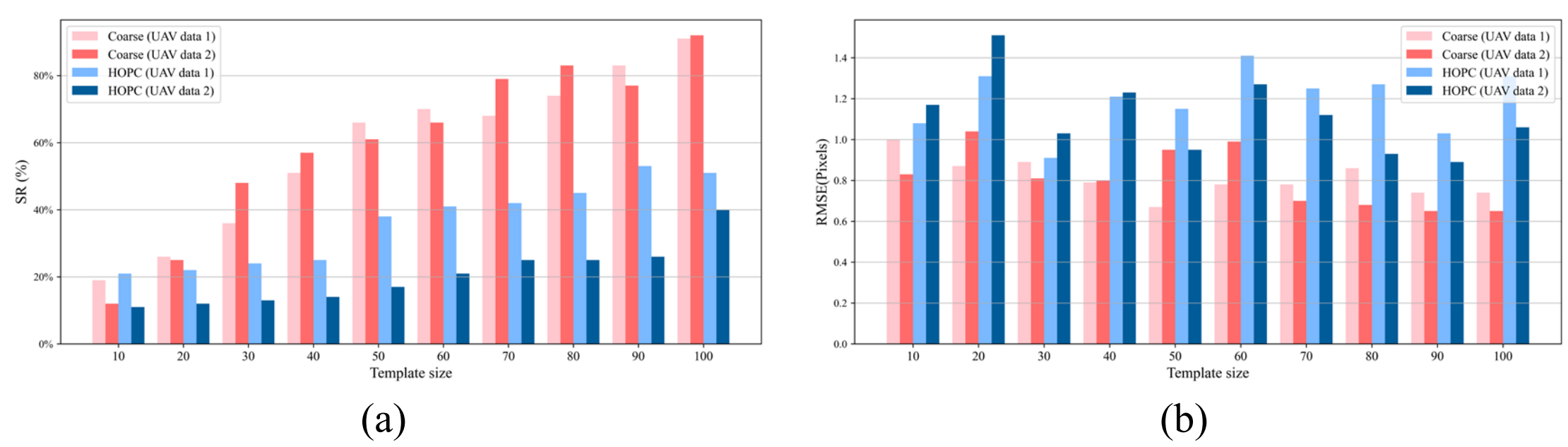

(a) (b)

Figure 5. Comparison of the two methods with different template size. (a) SR result. (b) RMSE results.

Regarding the RMSE metric, the influence of window size on both methods is evident. A larger window generally improves the accuracy of multimodal matching. However, due to the effect of PC noise processing, the RMSE of the HOPC method remains around 1 pixel, whereas that of our method is approximately 0.8 pixels, as summarized in Table II. As shown in Figure 5(b), our method consistently achieves lower RMSE values than the HOPC method because it accounts for the structural impact of PC noise.

Table 2. statistical comparison of the average values of various indicators of the two methods.

| Methods | NCM | CMR | SR | RMSE (pixels) |
|---|---|---|---|---|
| HOPC | 56 | 91% | 28% | 1.12 |
| Coarse matching | **120** | 92% | **59%** | **0.81** |

### *4.4. Performance evaluation of fine matching*

To test the performance of fine matching, we compared PC with the absolute minimum bias strategy (PCLAD) without mutual-structure, the commonly used least squares matching (PCLSM), and our proposed PCWLAD. Based on coarse matching results with different window sizes, the average coarse matching NCM is 730. We tested the NCM, CMR, SR, and RMSE metrics, as well as the iterative convergence performance.

Among the four metrics—NCM, CMR, SR, and RMSE—our method achieves the best performance in NCM, SR, and RMSE, as reflected by the average values in the table 3. Figures Figure6 (a) and Figure6 (b) illustrate the variations in SR and RMSE with respect to window size for all three methods, showing that larger windows generally yield better performance. Compared with PCLAD, our method attains higher SR and lower RMSE because the cross-structure weighting effectively suppresses noise and compensates for inconsistent structural textures. In contrast, PCLSM is more sensitive to noise and often fails to converge during iteration. The comparable CMR values of PCWLAD and PCLAD result from both methods already achieving relatively good coarse matching performance.

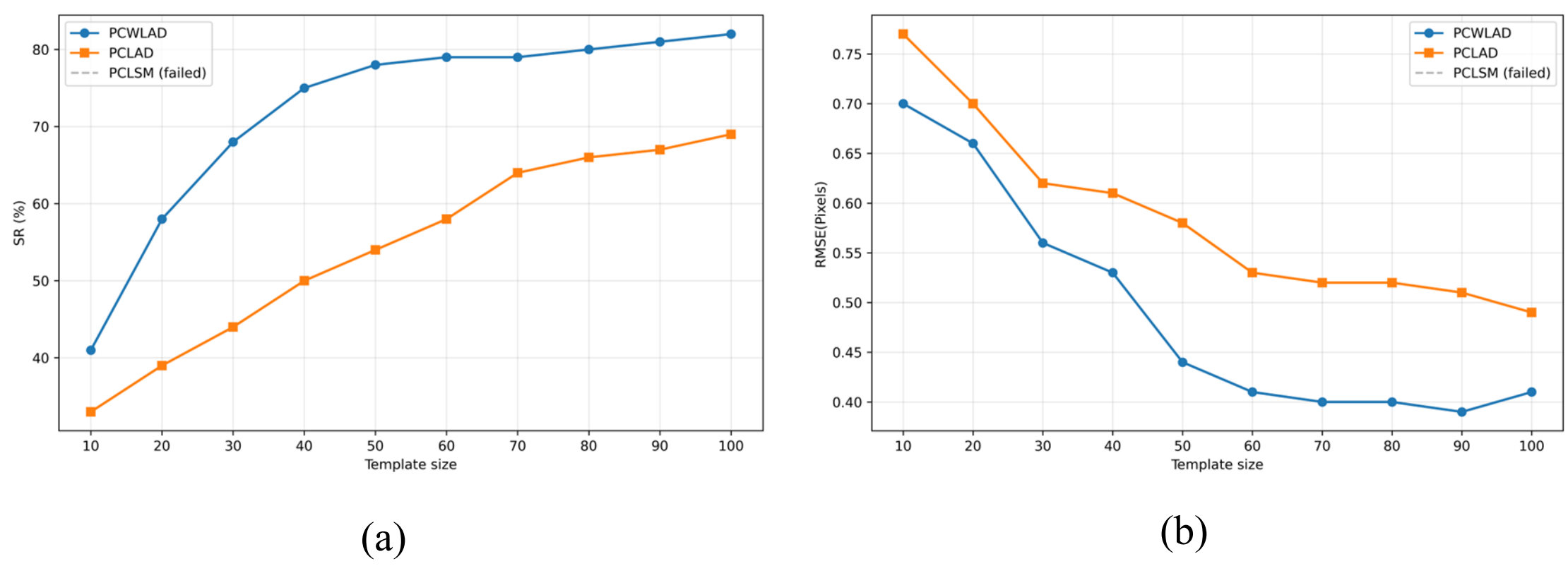

(a) (b)

Figure 6. Fine matching results for different template sizes. (a) SR results. (b) RMSE results.

Table 3. Statistical comparison of the average values of various indicators using the three methods.

| Method | NCM | CMR | SR | RMSE (pixels) |
| --- | --- | --- | --- | --- |
| PCWLAD | **526** | 99% | **72%** | **0.49** |
| PCLAD | 394 | 99% | **54%** | 0.59 |
| PCLSM | - | - | - | - |

### *4.5. Qualitative results and analysis*

The process is compared against eight state-of-the-art approaches, including area-based methods (HOPC and CFOG), feature-based methods (RIFT, HAPCG, and OFM), and deep learning-based methods (SuperGlue, LoFTR, and RoMa). All comparison methods use the default parameters.

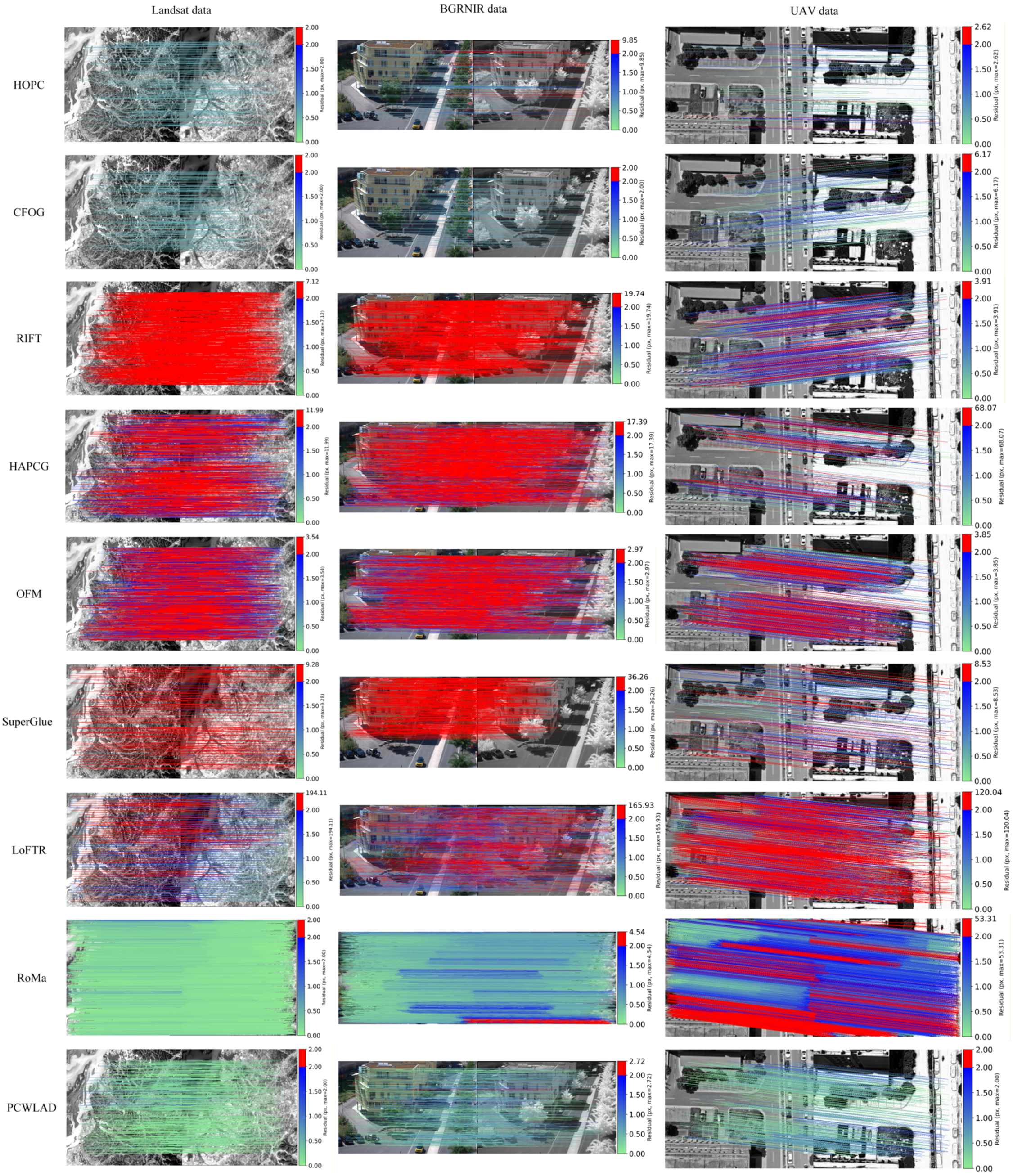

Figure 7. The figure illustrates the matching lines for each method, with different colors indicating the magnitude of the residuals after evaluating the matching points. Residuals between 0 and 2 pixels are considered inliers, while those exceeding 2 pixels are outliers, shown as red lines.

Figure 7 shows the matching results across three image pairs. Area-based methods, HOPC and CFOG, have few matching outliers and residuals around one pixel, but their matching points are limited, especially in UAV data, likely due to viewpoint-induced deformation affecting their SR. Feature-based methods like RIFT have many outliers and often lack sub-pixel accuracy, primarily due to noise and structural inconsistencies in PC. Deep learning methods like SuperGlue and LoFTR also show many outliers, though LoFTR matches a small set of high-precision points. RoMa achieves dense, high-precision matching but has many outliers in BGRNIR and UAV data. In contrast, our PCWLAD method provides higher accuracy with fewer outliers and a significantly larger number of matching points than area-based methods.

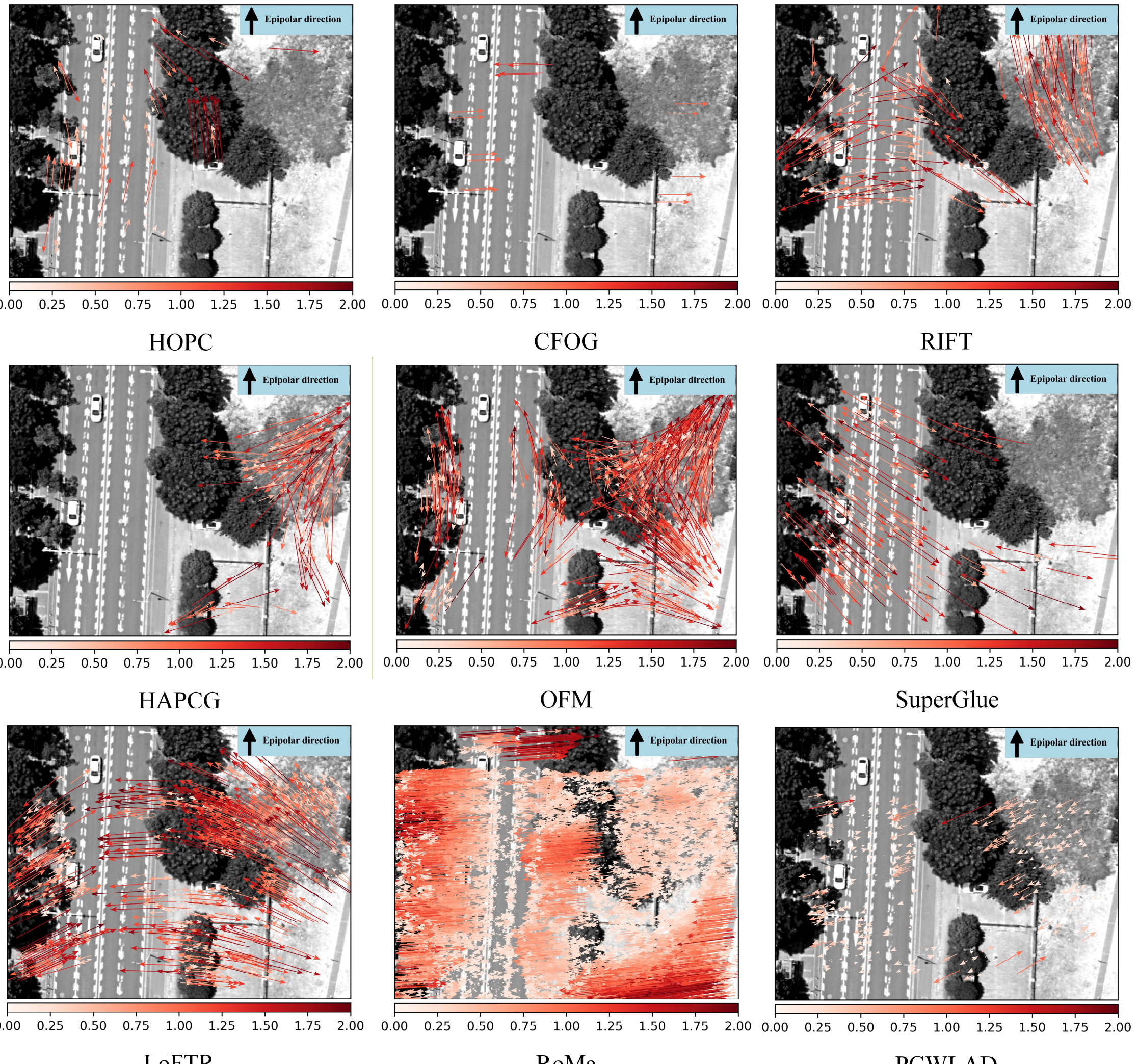

Figure 8. The distribution of the residuals of the reprojection of the fundamental matrix of the nine methods on UAV data 1. The direction of the red arrow indicates the residual direction and size under the epipolar line constraint, and the color intensity indicates the residual size. The epipolar line direction is roughly consistent with the baseline direction.

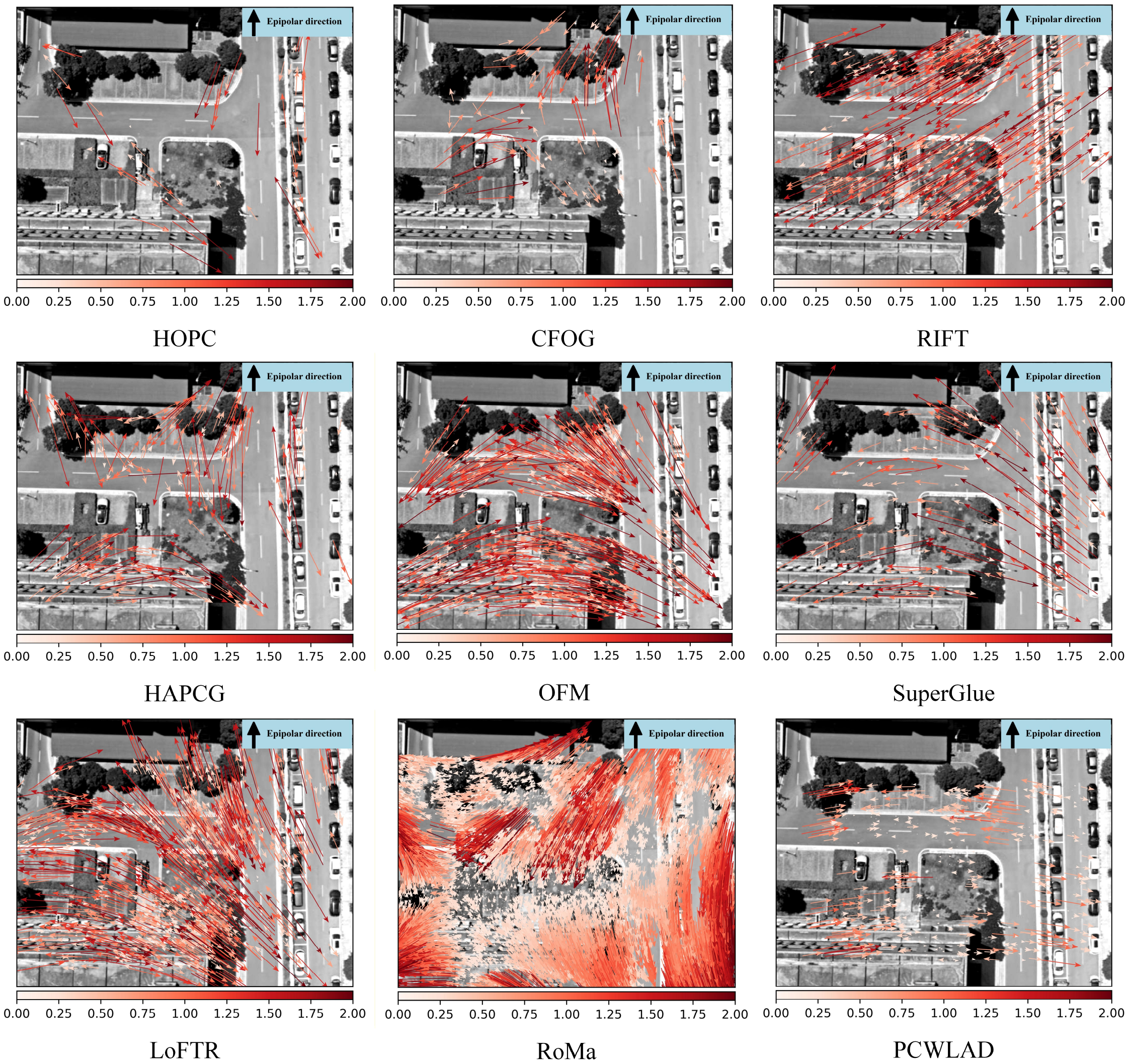

Figure 9. The distribution of the residuals of the reprojection of the fundamental matrix of the nine methods on UAV data 2.

We use two pairs of images with different perspectives to compute the fundamental matrix and evaluate accuracy based on the reprojection residuals, as shown in Figures 8 and 9. The two aerial images captured by the UAV conform to epipolar geometry. After relative orientation, the coordinate differences of corresponding points in the two images should obey the epipolar geometry, with residuals perpendicular to the epipolar line direction. The residuals are perpendicular to the baseline direction when the two images are taken from orthogonal views. Therefore, the residuals can be used to evaluate the matching accuracy, as illustrated in Figure 8. Under flat terrain conditions, it can be observed that CFOG, RoMa, and PCWLAD maintain residuals roughly perpendicular to the epipolar lines, indicating good internal consistency. In contrast, other methods fail to meet this constraint. In Figure 9, under conditions of significant perspective changes and undulating terrain, only the proposed PCWLAD method maintains residuals perpendicular to the epipolar lines, further demonstrating the superior internal consistency of our approach. The main reason is that when the image has complex local deformation, methods such as Log-Gabor filtering use the same parameters for filtering, and the structural noise generated by their response is more likely to propagate to the PC image.

### *4.6. Quantitative indicator comparison and analysis*

We evaluated nine methods on six image pairs using three metrics: CMR, RMSE, and NCM. For RMSE calculation, only matches with residuals under 2 pixels were considered valid. The results in Figure 10 show that the proposed PCWLAD method outperforms all other methods across nearly all datasets. The CMR and matching accuracy are significantly improved by noise suppression and optimization strategies applied to the PC maps.

As shown in Figure 10(a), the CMR performance of the proposed PCWLAD method remains remarkably stable, close to 100% across all datasets with minimal fluctuations. RoMa, HOPC, and CFOG achieve high, stable CMR values. LoFTR's median CMR is slightly lower than HOPC but still outperforms OFM and HAPCG. The CMR distributions of OFM and HAPCG are moderate, typically between 60% and 80%, indicating stable performance. In contrast, RIFT and SuperGlue show significantly lower CMR values, with occasional matching failures. A comprehensive comparison of the average results for all three image data categories is provided in Table 4.

Table 4 and Figure 10(b) show that the proposed PCWLAD method achieves the highest accuracy among all evaluated methods, with the smallest and most compactly distributed RMSE values, averaging approximately 0.34 pixels across all datasets. RoMa's average RMSE is around 0.4 pixels, slightly better than PCWLAD on the Landsat and BGRNIR datasets. However, on real UAV data, RoMa's RMSE is about 0.3 pixels higher than PCWLAD's. CFOG, HOPC, and LoFTR also yield low, stable RMSEs ranging from 0.6 to 0.9 pixels. In contrast, OFM and HAPCG have notably higher RMSEs, exceeding 1 pixel. SuperGlue and RIFT perform the worst, showing the highest RMSEs and significant fluctuations across datasets.

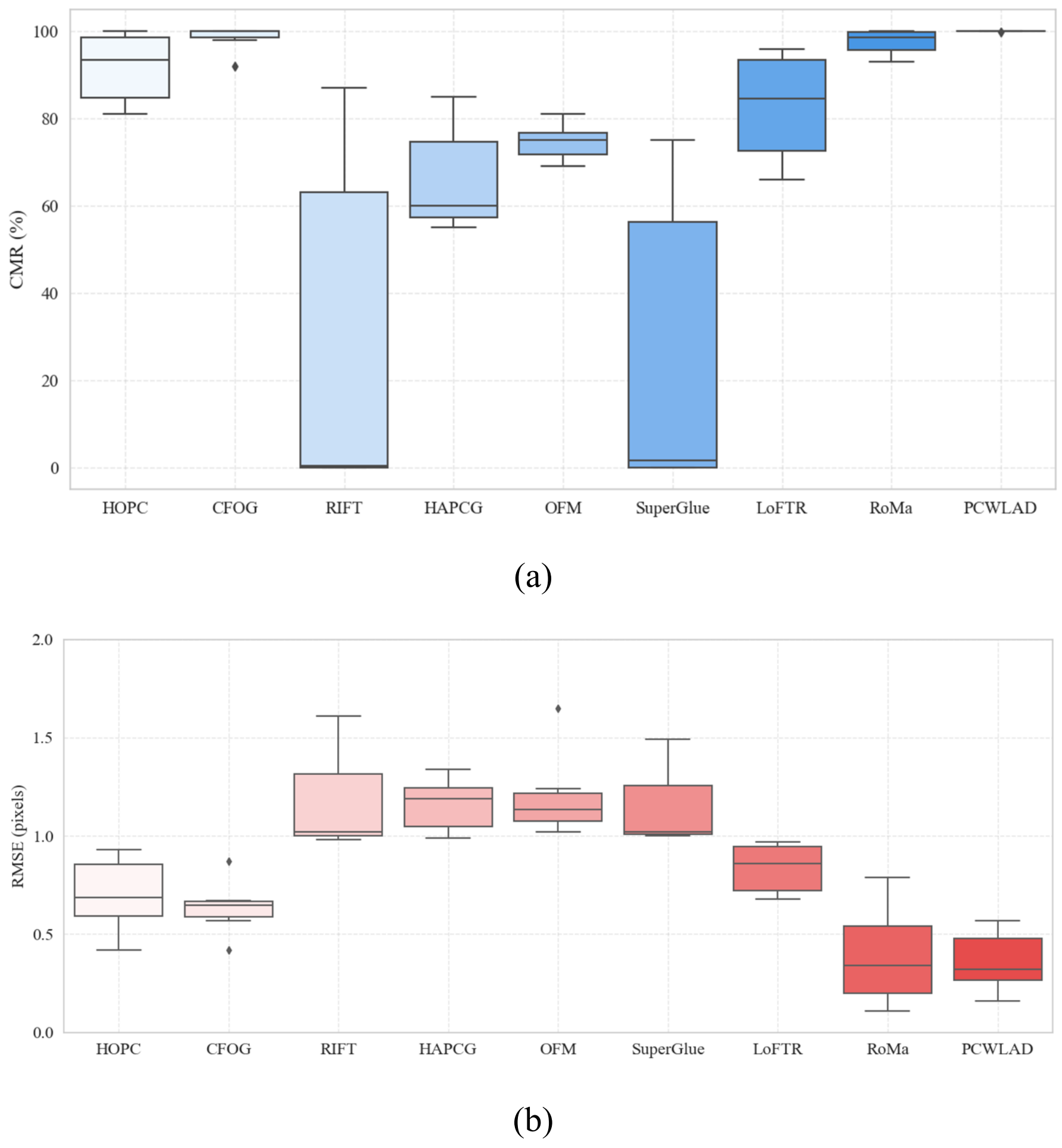

(a)

(b)

Figure 10. The box plot provides a quantitative comparison of different metrics across various datasets. (a) CMR results. (b) RMSE results.

Regarding NCM, RoMa matched a dense set of points, while LoFTR, OFM, and HAPCG also achieved relatively high values. Due to their low CMR, RIFT and SuperGlue produced the fewest correct matches. HOPC and CFOG matched about 140 points on average. In contrast, the proposed PCWLAD method achieved around 610 correct matches, significantly more than area-based methods like HOPC and CFOG, despite extracting only 1,000 features. Our software implementation allows users to adjust the number of extracted features as a configurable input parameter to suit different application needs.

Table 4. Test results of different methods on various datasets.

| Data | Landsat data | | | BGRNIR data | | | UAV data | | |
|---|---|---|---|---|---|---|---|---|---|
| | NCM | CMR | RMSE | NCM | CMR | RMSE | NCM | CMR | RMSE |
| HOPC | 200 | 100% | 0.52 | 133 | 81% | 0.68 | 77 | 93% | 0.91 |
| CFOG | 199 | 100% | 0.51 | 160 | 96% | 0.65 | 84 | 98% | 0.77 |
| RIFT | 5 | 0.4% | 1.61 | 0 | 0 | - | 391 | 85% | 1.00 |
| HAPCG | 1019 | 57% | 1.30 | 1870 | 58% | 1.19 | 206 | 80% | 1.00 |
| OFM | 1641 | 72% | 1.44 | 3215 | 73% | 1.13 | 665 | 78% | 1.05 |
| SuperGlue | 4 | 0.2% | 1.49 | 0 | 0 | - | 172 | 75% | 1.01 |
| LoFTR | 3062 | 95% | 0.82 | 5427 | 69% | 0.74 | 717 | 69% | 0.96 |
| RoMa | 10000 | 100% | **0.16** | 9700 | 97% | 0.55 | 9550 | 95% | 0.70 |
| PCWLAD | 800 | **100%** | **0.21** | 607 | **99%** | **0.43** | 413 | **100%** | **0.4** |

Quantitative results show that area-based methods like HOPC and CFOG effectively match optical multimodal images with nonlinear intensity differences. However, they are limited by the small number of matching points and integer pixel accuracy, making them unsuitable for high-precision geometric calibration. Feature-based methods like OFM and HAPCG can match more points but often require outlier removal, and their accuracy is insufficient for precise geometric processing. These methods are also affected by PC noise and structural inconsistencies, leading to pixel-level biases. Deep learning-based methods improve matching robustness. RoMa achieves high accuracy and many matching points on Landsat data but performs poorly with viewpoint variations. The PCWLAD method, by integrating noise features and structural information from the PC maps, mitigates noise interference, enabling precise sub-pixel offset recovery.

## 5.Conclusions

This study proposes a sub-pixel matching method for optical multimodal images, called PCWLAD. The PCWLAD framework consists of two main steps: coarse matching and fine matching. A PC map without noise suppression is used in the coarse matching stage, and the SSIM coefficient serves as the matching criterion. For precision optimization, sub-pixel displacement is accurately estimated through mutual structure weighting and the WLAD criterion. The method is evaluated on three types of multimodal optical images. PCWLAD demonstrates significant advantages in both CMR and RMSE, and outperforms eight state-of-the-art methods regarding internal consistency and matching accuracy.

PCWLAD is less effective when dealing with large geometric deformations. Fortunately, remote sensing images typically come with accurate georeferencing, which can be used for rough alignment to alleviate this issue. A more effective solution is integrating PCWLAD with feature-based matching methods, using the initial matches to provide better orientation and offset estimates. This reduces the search space and improves robustness to deformation, enabling the process to handle geometric and radiometric variations while maintaining high-precision matching.

## Disclosure statement

No potential conflict of interest was reported by the author(s).

## Funding

This study was supported by the National Natural Science Foundation of China (NSFC) project [grant numbers 42271410].

## References

Aggarwal, Alankrita, Mamta Mittal, and Gopi Battineni. 2021. "Generative adversarial network: An overview of theory and applications." *International Journal of Information Management Data Insights* 1 (1):100004.

Barath, D., D. Mishkin, M. Polic, W. Förstner, and J. Matas. 2023. A Large-Scale Homography Benchmark. Paper presented at the 2023 IEEE/CVF Conference on Computer Vision and Pattern Recognition (CVPR), 17-24 June 2023.

Brown, M., and S. Süsstrunk. 2011. Multi-spectral SIFT for scene category recognition. Paper presented at the CVPR 2011, 20-25 June 2011.

Chang, H. H., G. L. Wu, and M. H. Chiang. 2019. "Remote Sensing Image Registration Based on Modified SIFT and Feature Slope Grouping." *IEEE Geoscience and Remote Sensing Letters* 16 (9):1363-1367. https://doi.org/10.1109/LGRS.2019.2899123.

Cui, S., A. Ma, L. Zhang, M. Xu, and Y. Zhong. 2022. "MAP-Net: SAR and Optical Image Matching via Image-Based Convolutional Network With Attention Mechanism and Spatial Pyramid Aggregated Pooling." *IEEE Transactions on Geoscience and Remote Sensing* 60:1-13. https://doi.org/10.1109/TGRS.2021.3066432.

Dalal, N., and B. Triggs. 2005. Histograms of oriented gradients for human detection. Paper presented at the 2005 IEEE Computer Society Conference on Computer Vision and Pattern Recognition (CVPR'05), 20-25 June 2005.

Daubechies, Ingrid, Ronald DeVore, Massimo Fornasier, and C. Sinan Güntürk. 2010. "Iteratively reweighted least squares minimization for sparse recovery." *Communications on Pure and Applied Mathematics* 63 (1):1-38. https://doi.org/https://doi.org/10.1002/cpa.20303.

Donoho, D. L. 1995. "De-noising by soft-thresholding." *IEEE Transactions on Information Theory* 41 (3):613-627. https://doi.org/10.1109/18.382009.

Edstedt, J., Q. Sun, G. Bökman, M. Wadenbäck, and M. Felsberg. 2024. RoMa: Robust Dense Feature Matching. Paper presented at the 2024 IEEE/CVF Conference on Computer Vision and Pattern Recognition (CVPR), 16-22 June 2024.

Fan, Z., M. Wang, Y. Pi, Y. Liu, and H. Jiang. 2023. "A Robust Oriented Filter-Based Matching Method for Multisource, Multitemporal Remote Sensing Images." *IEEE Transactions on Geoscience and Remote Sensing* 61:1-16. https://doi.org/10.1109/TGRS.2023.3288531.

Fan, Zhongli, Yuxian Liu, Yuxuan Liu, Li Zhang, Junjun Zhang, Yushan Sun, and Haibin Ai. 2022. "3MRS: an effective coarse-to-fine matching method for multimodal remote sensing imagery." *Remote Sensing* 14 (3):478.

Fan, Zhongli, Yingdong Pi, and Mi Wang. 2025. "Effective feature matching for multimodal remote sensing images via sparse sampling description." *ISPRS Journal of Photogrammetry and Remote Sensing* 230:943-961. https://doi.org/https://doi.org/10.1016/j.isprsjprs.2025.10.012.

Fischler, Martin A., and Robert C. Bolles. 1981. "Random sample consensus: a paradigm for model fitting with applications to image analysis and automated cartography." *Commun. ACM* 24:381-395.

Gruen, Armin. 1985. "Adaptive least squares correlation: a powerful image matching technique." *South African Journal of Photogrammetry, Remote Sensing and Cartography* 14 (3):175-187.

Harris, Chris, and Mike Stephens. 1988. A combined corner and edge detector. Paper presented at the Alvey vision conference.

He, Yunze, Baoyuan Deng, Hongjin Wang, Liang Cheng, Ke Zhou, Siyuan Cai, and Francesco Ciampa. 2021. "Infrared machine vision and infrared thermography with deep learning: A review." *Infrared Physics & Technology* 116:103754. https://doi.org/https://doi.org/10.1016/j.infrared.2021.103754.

Hou, Zhuolu, Yuxuan Liu, and Li Zhang. 2024. "POS-GIFT: A geometric and intensity-invariant feature transformation for multimodal images." *Information Fusion* 102:102027.

Hu, Zhuoyue, Xiaoyan Li, Liyuan Li, Xiaofeng Su, Lin Yang, Yong Zhang, Xingjian Hu, et al. 2024. "Wide-swath and high-resolution whisk-broom imaging and on-orbit performance of SDGSAT-1 thermal infrared spectrometer." *Remote Sensing of Environment* 300:113887. https://doi.org/https://doi.org/10.1016/j.rse.2023.113887.

Huang, T., H. Pan, and N. Zhou. 2024. "Adaptive parameter local consistency automatic outlier removal algorithm for area-based matching." *ISPRS Ann. Photogramm. Remote Sens. Spatial Inf. Sci.* X-1-2024:99-106. https://doi.org/10.5194/isprs-annals-X-1-2024-99-2024.

Huang, Wei, Tianrui Li, Jia Liu, Peng Xie, Shengdong Du, and Fei Teng. 2021. "An overview of air quality analysis by big data techniques: Monitoring, forecasting, and traceability." *Information Fusion* 75:28-40. https://doi.org/https://doi.org/10.1016/j.inffus.2021.03.010.

Hughes, Lloyd Haydn, Diego Marcos, Sylvain Lobry, Devis Tuia, and Michael Schmitt. 2020. "A deep learning framework for matching of SAR and optical imagery." *ISPRS Journal of Photogrammetry and Remote Sensing* 169:166-179. https://doi.org/https://doi.org/10.1016/j.isprsjprs.2020.09.012.

Jiang, Chenchen, Huazhong Ren, Hong Yang, Hongtao Huo, Pengfei Zhu, Zhaoyuan Yao, Jing Li, Min Sun, and Shihao Yang. 2024. "M2FNet: Multi-modal fusion network for object detection from visible and thermal infrared images." *International Journal of Applied Earth Observation and Geoinformation* 130:103918. https://doi.org/https://doi.org/10.1016/j.jag.2024.103918.

Jiang, Xingyu, Jiayi Ma, Guobao Xiao, Zhenfeng Shao, and Xiaojie Guo. 2021. "A review of multimodal image matching: Methods and applications." *Information Fusion* 73:22-71. https://doi.org/https://doi.org/10.1016/j.inffus.2021.02.012.

Kim, Bongjoe, Jihoon Choi, Yongwoon Park, and Kwanghoon Sohn. 2012. "Robust Corner Detection Based on Image Structure." *Circuits, Systems, and Signal Processing* 31 (4):1443-1457. https://doi.org/10.1007/s00034-012-9388-z.

Kovesi, Peter. 1999. "Image features from phase congruency." *Videre: Journal of computer vision research* 1 (3):1-26.

Li, Jiayuan, Qingwu Hu, and Mingyao Ai. 2019. "RIFT: Multi-Modal Image Matching Based on Radiation-Variation Insensitive Feature Transform." *IEEE Transactions on Image Processing* PP:1-1. https://doi.org/10.1109/TIP.2019.2959244.

Li, Liangzhi, Ling Han, and Yuanxin Ye. 2022. "Self-Supervised Keypoint Detection and Cross-Fusion Matching Networks for Multimodal Remote Sensing Image Registration." *Remote Sensing* 14 (15):3599. https://doi.org/10.3390/rs14153599.

Li, Liangzhi, Ling Han, Yuanxin Ye, Yuming Xiang, and Tingyu Zhang. 2025. "Deep learning in remote sensing image matching: A survey." *ISPRS Journal of Photogrammetry and Remote Sensing* 225:88-112. https://doi.org/https://doi.org/10.1016/j.isprsjprs.2025.04.001.

Liang, Chenbin, Yunyun Dong, Changjun Zhao, and Zengguo Sun. 2023. "A Coarse-to-Fine Feature Match Network Using Transformers for Remote Sensing Image Registration." *Remote Sensing* 15 (13):3243. https://doi.org/10.3390/rs15133243.

Liao, Yifan, Ke Xi, Huijin Fu, Lai Wei, Shuo Li, Qiang Xiong, Qi Chen, Pengjie Tao, and Tao Ke. 2024. "Refining multi-modal remote sensing image matching with repetitive feature optimization." *International Journal of Applied Earth Observation and Geoinformation* 134:104186. https://doi.org/https://doi.org/10.1016/j.jag.2024.104186.

Lindenberger, P., P. E. Sarlin, and M. Pollefeys. 2023. LightGlue: Local Feature Matching at Light Speed. Paper presented at the 2023 IEEE/CVF International Conference on Computer Vision (ICCV), 1-6 Oct. 2023.

Liu, S., and W. Deng. 2015. Very deep convolutional neural network based image classification using small training sample size. Paper presented at the 2015 3rd IAPR Asian Conference on Pattern Recognition (ACPR), 3-6 Nov. 2015.

Lowe, David G. 2004. "Distinctive Image Features from Scale-Invariant Keypoints." *International Journal of Computer Vision* 60 (2):91-110. https://doi.org/10.1023/B:VISI.0000029664.99615.94.

Ma, Jiayi, Yong Ma, and Chang Li. 2019. "Infrared and visible image fusion methods and applications: A survey." *Information Fusion* 45:153-178. https://doi.org/https://doi.org/10.1016/j.inffus.2018.02.004.

Maes, F., A. Collignon, D. Vandermeulen, G. Marchal, and P. Suetens. 1997. "Multimodality image registration by maximization of mutual information." *IEEE transactions on medical imaging* 16 (2):187-198. https://doi.org/10.1109/42.563664.

Maimaitijiang, Maitiniyazi, Vasit Sagan, Paheding Sidike, Sean Hartling, Flavio Esposito, and Felix B. Fritschi. 2020. "Soybean yield prediction from UAV using multimodal data fusion and deep learning." *Remote Sensing of Environment* 237:111599. https://doi.org/https://doi.org/10.1016/j.rse.2019.111599.

Meng, Lingxuan, Ji Zhou, Shaomin Liu, Ziwei Wang, Xiaodong Zhang, Lirong Ding, Li Shen, and Shaofei Wang. 2022. "A robust registration method for UAV thermal infrared and visible images taken by dual-cameras." *ISPRS Journal of Photogrammetry and Remote Sensing* 192:189-214. https://doi.org/https://doi.org/10.1016/j.isprsjprs.2022.08.018.

Merkle, Nina, Wenjie Luo, Stefan Auer, Rupert Müller, and Raquel Urtasun. 2017. "Exploiting deep matching and SAR data for the geo-localization accuracy improvement of optical satellite images." *Remote Sensing* 9 (6):586.

Motayyeb, Soroush, Farhad Samadzedegan, Farzaneh Dadrass Javan, and Hamidreza Hosseinpour. 2023. "Fusion of UAV-based infrared and visible images for thermal leakage map generation of building facades." *Heliyon* 9 (3).

Nguyen, T., S. W. Chen, S. S. Shivakumar, C. J. Taylor, and V. Kumar. 2018. "Unsupervised Deep Homography: A Fast and Robust Homography Estimation Model." *IEEE Robotics and Automation Letters* 3 (3):2346-2353. https://doi.org/10.1109/LRA.2018.2809549.

Nunes, C. F. G., and F. L. C. Pádua. 2024. "An Orientation-Robust Local Feature Descriptor Based on Texture and Phase Congruency for Visible–Infrared Image Matching." *IEEE Geoscience and Remote Sensing Letters* 21:1-5. https://doi.org/10.1109/LGRS.2024.3379091.

Oquab, Maxime, Timothée Darcet, Théo Moutakanni, Huy Vo, Marc Szafraniec, Vasil Khalidov, Pierre Fernandez, Daniel Haziza, Francisco Massa, and Alaaeldin El-Nouby. 2023. "Dinov2: Learning robust visual features without supervision." *arXiv preprint arXiv:2304.07193*.

Quan, Dou, Shuang Wang, Yu Gu, Ruiqi Lei, Bowu Yang, Shaowei Wei, Biao Hou, and Licheng Jiao. 2022. "Deep feature correlation learning for multi-modal remote sensing image registration." *IEEE Transactions on Geoscience and Remote Sensing* 60:1-16.

Revaud, Jerome, Philippe Weinzaepfel, Zaid Harchaoui, and Cordelia Schmid. 2016. "DeepMatching: Hierarchical Deformable Dense Matching." *International Journal of Computer Vision* 120 (3):300-323. https://doi.org/10.1007/s11263-016-0908-3.

Sarlin, Paul-Edouard, Daniel DeTone, Tomasz Malisiewicz, and Andrew Rabinovich. 2020. Superglue: Learning feature matching with graph neural networks. Paper presented at the Proceedings of the IEEE/CVF conference on computer vision and pattern recognition.

Schönberger, J. L., and J. M. Frahm. 2016. Structure-from-Motion Revisited. Paper presented at the 2016 IEEE Conference on Computer Vision and Pattern Recognition (CVPR), 27-30 June 2016.

Sedaghat, Amin, and Hamid Ebadi. 2015. "Accurate Affine Invariant Image Matching Using Oriented Least Square." *Photogrammetric Engineering & Remote Sensing* 81 (9):733-743. https://doi.org/10.14358/pers.81.9.733.

Shechtman, E., and M. Irani. 2007. Matching Local Self-Similarities across Images and Videos. Paper presented at the 2007 IEEE Conference on Computer Vision and Pattern Recognition, 17-22 June 2007.

Shen, X., C. Zhou, L. Xu, and J. Jia. 2015. Mutual-Structure for Joint Filtering. Paper presented at the 2015 IEEE International Conference on Computer Vision (ICCV), 7-13 Dec. 2015.

Shin, Dongjoe, and Jan-Peter Muller. 2012. "Progressively weighted affine adaptive correlation matching for quasi-dense 3D reconstruction." *Pattern Recognition* 45 (10):3795-3809. https://doi.org/10.1016/j.patcog.2012.03.023.

Sun, Jiaming, Zehong Shen, Yuang Wang, Hujun Bao, and Xiaowei Zhou. 2021. LoFTR: Detector-free local feature matching with transformers. Paper presented at the Proceedings of the IEEE/CVF conference on computer vision and pattern recognition.

Suri, S., and P. Reinartz. 2010. "Mutual-Information-Based Registration of TerraSAR-X and Ikonos Imagery in Urban Areas." *IEEE Transactions on Geoscience and Remote Sensing* 48 (2):939-949. https://doi.org/10.1109/TGRS.2009.2034842.

Tilton, J. C., G. Lin, and B. Tan. 2017. "Measurement of the Band-to-Band Registration of the SNPP VIIRS Imaging System From On-Orbit Data." *IEEE Journal of Selected Topics in Applied Earth Observations and Remote Sensing* 10 (3):1056-1067. https://doi.org/10.1109/JSTARS.2016.2601561.

Tilton, J. C., R. E. Wolfe, G. Lin, and J. J. Dellomo. 2019. "On-Orbit Measurement of the Effective Focal Length and Band-to-Band Registration of Satellite-Borne Whiskbroom Imaging Sensors." *IEEE Journal of Selected Topics in Applied Earth Observations and Remote Sensing* 12 (11):4622-4633. https://doi.org/10.1109/JSTARS.2019.2949677.

Uss, Mykhail, Benoit Vozel, Vladimir Lukin, and Kacem Chehdi. 2022. "Exhaustive search of correspondences between multimodal remote sensing images using convolutional neural network." *Sensors* 22 (3):1231.

Viswanathan, Deepak Geetha. 2009. Features from accelerated segment test (fast). Paper presented at the Proceedings of the 10th workshop on image analysis for multimedia interactive services, London, UK.

Wang, Hansheng, Guodong Li, and Guohua Jiang. 2007. "Robust Regression Shrinkage and Consistent Variable Selection Through the LAD-Lasso." *Journal of Business & Economic Statistics* 25 (3):347-355. https://doi.org/10.1198/073500106000000251.

Wang, Z., and A. C. Bovik. 2009. "Mean squared error: Love it or leave it? A new look at Signal Fidelity Measures." *IEEE Signal Processing Magazine* 26 (1):98-117. https://doi.org/10.1109/MSP.2008.930649.

Wulder, Michael A., David P. Roy, Volker C. Radeloff, Thomas R. Loveland, Martha C. Anderson, David M. Johnson, Sean Healey, et al. 2022. "Fifty years of Landsat science and impacts." *Remote Sensing of Environment* 280:113195. https://doi.org/https://doi.org/10.1016/j.rse.2022.113195.

Xiang, Yuming, Feng Wang, and Hongjian You. 2018. "OS-SIFT: A robust SIFT-like algorithm for high-resolution optical-to-SAR image registration in suburban areas." *IEEE Transactions on Geoscience and Remote Sensing* 56 (6):3078-3090.

Yang, Z., T. Dan, and Y. Yang. 2018. "Multi-Temporal Remote Sensing Image Registration Using Deep Convolutional Features." *IEEE Access* 6:38544-38555. https://doi.org/10.1109/ACCESS.2018.2853100.

Yao, Yongxiang, Yongjun Zhang, Yi Wan, Xinyi Liu, and Haoyu Guo. 2021. "Heterologous Images Matching Considering Anisotropic Weighted Moment and Absolute Phase Orientation." *Geomatics and Information Science of Wuhan University* 46 (11):1727-1736. https://doi.org/10.13203/j.whugis20200702.

Ye, F., Y. Su, H. Xiao, X. Zhao, and W. Min. 2018. "Remote Sensing Image Registration Using Convolutional Neural Network Features." *IEEE Geoscience and Remote Sensing Letters* 15 (2):232-236. https://doi.org/10.1109/LGRS.2017.2781741.

Ye, Y., L. Bruzzone, J. Shan, F. Bovolo, and Q. Zhu. 2019. "Fast and Robust Matching for Multimodal Remote Sensing Image Registration." *IEEE Transactions on Geoscience and Remote Sensing* 57 (11):9059-9070. https://doi.org/10.1109/TGRS.2019.2924684.

Ye, Yuanxin, Jie Shan, Lorenzo Bruzzone, and Li Shen. 2017. "Robust Registration of Multimodal Remote Sensing Images Based on Structural Similarity." *IEEE Transactions on Geoscience and Remote Sensing* 55 (5):2941-2958. https://doi.org/10.1109/tgrs.2017.2656380.

Ye, Zhipeng, Haoting Liu, Haiguang Li, Zhenhui Guo, Hui Zhao, Lin Zhang, Qiang Liu, and Qing Li. 2024. Visible-Infrared Images Matching Based on Deep Learning. Paper presented at the Man-Machine-Environment System Engineering, Singapore, 2024//.

Yoo, Jae-Chern, and Tae Hee Han. 2009. "Fast Normalized Cross-Correlation." *Circuits, Systems and Signal Processing* 28 (6):819-843. https://doi.org/10.1007/s00034-009-9130-7.

Yu, Qiuze, Dawen Ni, Yuxuan Jiang, Yuxuan Yan, Jiachun An, and Tao Sun. 2021. "Universal SAR and optical image registration via a novel SIFT framework based on nonlinear diffusion and a polar spatial-frequency descriptor." *ISPRS Journal of Photogrammetry and Remote Sensing* 171:1-17. https://doi.org/https://doi.org/10.1016/j.isprsjprs.2020.10.019.

Zhang, H., L. Lei, W. Ni, T. Tang, J. Wu, D. Xiang, and G. Kuang. 2022. "Optical and SAR Image Matching Using Pixelwise Deep Dense Features." *IEEE Geoscience and Remote Sensing Letters* 19:1-5. https://doi.org/10.1109/LGRS.2020.3039473.

Zhang, Xinyue, Chengcai Leng, Yameng Hong, Zhao Pei, Irene Cheng, and Anup Basu. 2021. "Multimodal Remote Sensing Image Registration Methods and Advancements: A Survey." *Remote Sensing* 13 (24):5128.

Zhao, Liping, Xianhui Dou, Fan Mo, Hongmo Li, Fangxu Zhang, Dian Qu, and Junfeng Xie. 2024. "Geometric accuracy evaluation and analysis of ZY-1 02E IRS thermal infrared image data using GCP extraction based on phase correlation matching method." *The International Archives of the Photogrammetry, Remote Sensing and Spatial Information Sciences* 48:895-901.

Zhao, Yanhua, and Yan Li. 2021. "Technical Characteristics and Application of Visible and Infrared Multispectral Imager." In, 197-208.

Zheng, C., S. Li, C. Wang, and B. Zhang. 2025. "MSG: Robust Multimodal Remote Sensing Image Matching Using Side Window Gaussian Space." *IEEE Transactions*

*on Geoscience and Remote Sensing* 63:1-23. https://doi.org/10.1109/TGRS.2025.3591504.
Zitová, Barbara, and Jan Flusser. 2003. "Image registration methods: a survey." *Image and Vision Computing* 21 (11):977-1000. https://doi.org/https://doi.org/10.1016/S0262-8856(03)00137-9.